\documentclass{article}

\usepackage{microtype}
\usepackage{graphicx}
\usepackage{subcaption}
\usepackage{booktabs} 

\usepackage{hyperref}

\usepackage[preprint]{icml2026}

\usepackage{amsmath}
\usepackage{amssymb}
\usepackage{mathtools}
\usepackage{amsthm}

\usepackage[capitalize,noabbrev]{cleveref}

\usepackage{enumitem}
\usepackage[export]{adjustbox}
\usepackage{multirow}
\usepackage{xcolor}
\usepackage{microtype}
\usepackage{colortbl}
\usepackage{subcaption}

\newcommand{\vx}{\boldsymbol{x}}
\newcommand{\vz}{\boldsymbol{z}}
\newcommand{\vtheta}{\boldsymbol{\theta}}
\newcommand{\vp}{\boldsymbol{p}}
\newcommand{\vr}{\boldsymbol{r}}
\newcommand{\vv}{\boldsymbol{v}}
\newcommand{\veps}{\boldsymbol{\epsilon}}

\theoremstyle{plain}

\theoremstyle{definition}

\theoremstyle{remark}

\usepackage[textsize=tiny]{todonotes}

\icmltitlerunning{LLaDA-o: An Effective and Length-Adaptive Omni Diffusion Model}

\begin{document}

\twocolumn[
  \icmltitle{LLaDA-o: An Effective and Length-Adaptive Omni Diffusion Model}



  \icmlsetsymbol{intern}{$\dagger$}
  \icmlsetsymbol{cor}{$\mathparagraph$}

  \begin{icmlauthorlist}
    \icmlauthor{Zebin You}{ruc1,ruc2,ruc3,intern}
    \icmlauthor{Xiaolu Zhang}{ant}
    \icmlauthor{JUN ZHOU}{ant}
    \icmlauthor{Chongxuan Li}{ruc1,ruc2,ruc3,cor}
    \icmlauthor{Ji-Rong Wen}{ruc1,ruc2,ruc3}
  \end{icmlauthorlist}

  \icmlaffiliation{ruc1}{Gaoling School of Artificial Intelligence, Renmin University of China, Beijing, China.}
  \icmlaffiliation{ruc2}{Beijing Key Laboratory of Research on Large Models and Intelligent Governance.}
  \icmlaffiliation{ruc3}{Engineering Research Center of Next-Generation Intelligent Search and Recommendation, MOE.}
  \icmlaffiliation{ant}{Ant Group}
  \icmlcorrespondingauthor{Chongxuan Li}{chongxuanli@ruc.edu.cn}

  \icmlkeywords{Machine Learning, ICML}

  \vskip 0.3in
]



\printAffiliationsAndNotice{\icmlItern}

\begin{abstract}
We present \textbf{LLaDA-o}, an effective and length-adaptive omni diffusion model for multimodal understanding and generation. LLaDA-o is built on a Mixture of Diffusion (MoD) framework that decouples discrete masked diffusion for text understanding and continuous diffusion for visual generation, while coupling them through a shared, simple, and efficient attention backbone that reduces redundant computation for fixed conditions. Building on MoD, we further introduce a data-centric length adaptation strategy that enables flexible-length decoding in multimodal settings without architectural changes. Extensive experiments show that LLaDA-o achieves state-of-the-art performance among omni-diffusion models on multimodal understanding and generation benchmarks, and reaches 87.04 on DPG-Bench for text-to-image generation, supporting the effectiveness of unified omni diffusion modeling. Code is available at \url{https://github.com/ML-GSAI/LLaDA-o}.
\end{abstract}

\begin{figure*}[!t]
    \centering
\includegraphics[width=.8525\linewidth]{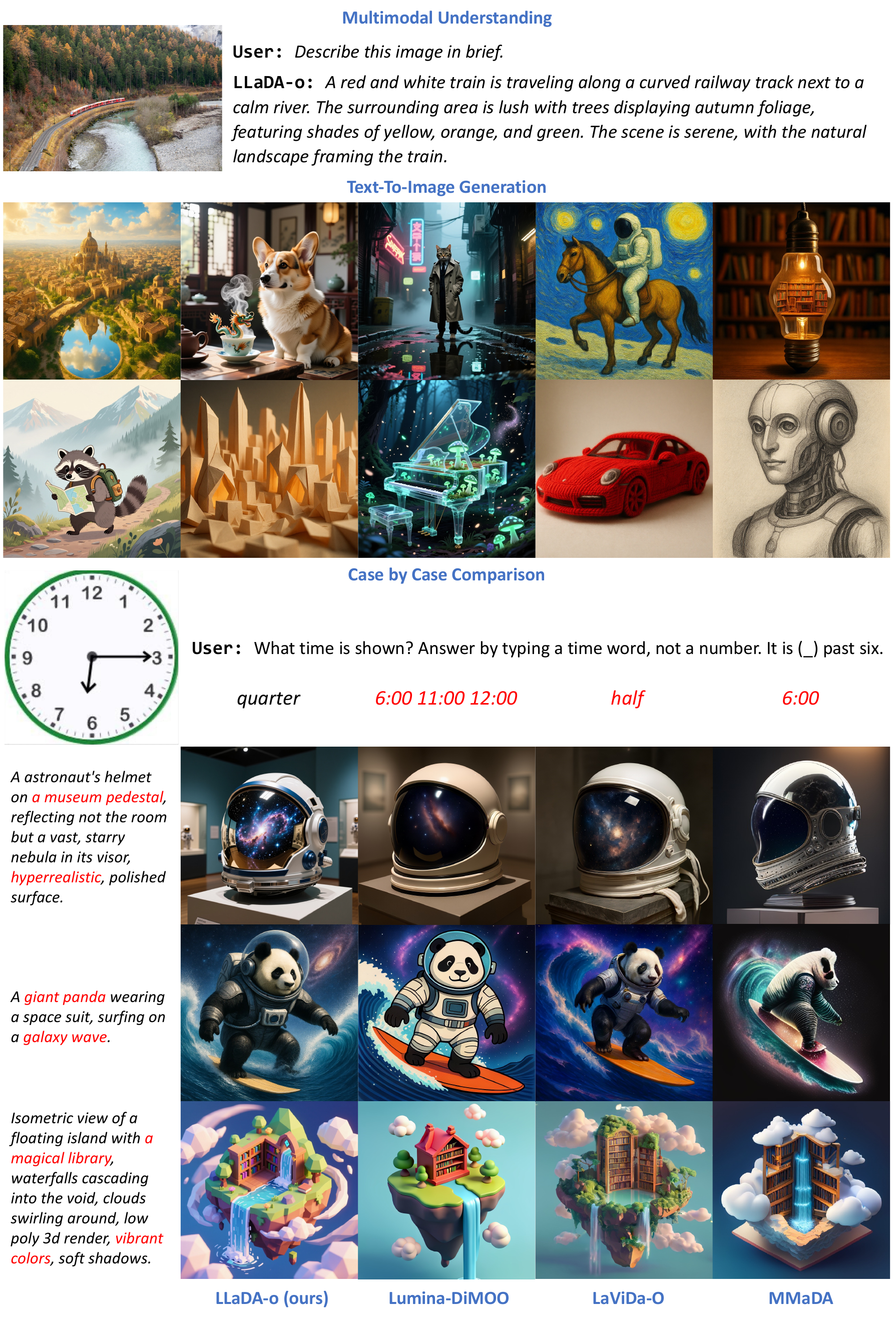}
\caption{\textbf{Overview of LLaDA-o's capabilities.} Top: multimodal understanding examples. Middle: text-to-image generation results following complex prompts (see Table~\ref{tab:image_prompts} for the prompts). Bottom: case by case comparison with existing omni diffusion models, where LLaDA-o achieves stronger understanding performance and generates images with richer fine-grained details following the instructions.}\label{fig:generated_image_demo}
\end{figure*}

\section{Introduction}
\label{sec:intro}

Masked diffusion models (MDMs)~\cite{austin2021structured,loudiscrete,shi2024simplified,sahoo2024simple,ou2024your} have recently emerged as a promising alternative to autoregressive (AR) language models. 
By iteratively denoising masked tokens in parallel, MDMs have demonstrated strong language modeling performance~\cite{nie2025large,ye2025dream,cheng2025sdar,bie2025llada2} and offer appealing properties such as bidirectional context modeling and improved inference parallelism~\cite{song2025seed,khanna2025mercury,gemini2025}, drawing increasing attention to diffusion-based language modeling.

Despite this progress, the potential of diffusion language models as \emph{omni} models for multimodal understanding and generation remains largely underexplored. 
A key challenge is that text and images favor fundamentally different diffusion dynamics: masked diffusion naturally operates over discrete language tokens, whereas for images, continuous diffusion in latent space has become the de facto standard~\cite{sohl2015deep,ho2020denoising,song2020score}. 
Although attempts exists~\cite{pynadath2025candi,chang2022maskgit}, these modality-specific preferences do not transfer trivially across domains, motivating a hybrid design that models text discretely and images continuously.
 
However, naively co-training both modalities within a single dense diffusion model is often ineffective. 
The heterogeneous state spaces and corruption processes can induce objective mismatch and gradient interference, leading to unstable optimization and suboptimal performance~\cite{li2025dual}. 
In addition, existing omni-diffusion models~\cite{xin2025lumina,li2025lavidao,yang2025mmada} often assume a fixed length for understanding, constraining their applicability open-ended settings.

To address these challenges, we propose \textbf{LLaDA-o}, an effective and length-adaptive omni diffusion model built upon the \emph{Mixture of Diffusion} (MoD) framework. 
MoD decouples modality-appropriate diffusion processes into specialized experts while maintaining a shared attention backbone for cross-modality interaction. 
Specifically, it assigns masked diffusion to an \emph{understanding expert} for text and visual encoder tokens, and continuous diffusion to a \emph{generation expert} for visual latent tokens, mitigating optimization conflicts in dense co-training. 
Building on MoD, we further introduce an efficient attention mechanism and a data-centric length adaptation strategy, enabling scalable inference and flexible-length generation in multimodal settings.

We evaluate LLaDA-o from both qualitative and quantitative perspectives. 
Qualitatively, Fig.~\ref{fig:generated_image_demo} shows that LLaDA-o achieves stronger multimodal understanding and produces images with richer fine-grained details than existing omni-diffusion models such as Lumina-DiMOO~\cite{xin2025lumina}. 
Quantitatively, we evaluate LLaDA-o on ten multimodal understanding benchmarks (see Tab.~\ref{tab:multimodal_understanding_results}), where it achieves state-of-the-art results among omni-diffusion models and other discrete-diffusion-based multimodal approaches. 
We further evaluate text-to-image generation on the widely used GenEval~\cite{ghosh2023geneval} and DPG-Bench~\cite{hu2024ella} benchmarks against strong generation-only and unified multimodal models, where LLaDA-o achieves \emph{state-of-the-art} performance on DPG-Bench (87.04). Besides, we present a comprehensive analysis for LLaDA-o. Overall, these results demonstrate the effectiveness of LLaDA-o as a unified omni diffusion model.

\section{Preliminaries}

We present preliminaries on diffusion models.

\subsection{Continuous Diffusion Models}

Continuous diffusion models (CDMs)~\cite{sohl2015deep,ho2020denoising,song2020score} constitute a core paradigm in modern image generation, particularly when combined with diffusion Transformers~\cite{peebles2023scalable,bao2023all}. 
Conceptually, a CDM specifies a forward stochastic process that gradually corrupts data into noise, and learns to reverse this process to generate samples from the noise distribution. This mechanism can be characterized using stochastic differential equations~\cite{song2020score}. At inference, the same dynamics admit an equivalent ordinary differential equation (ODE) formulation, leading to deterministic samplers~\cite{song2020denoising,lu2022dpm}. This ODE perspective has further inspired a series of variants, including rectified flow (RF)~\cite{liu2022rectified} and Flow Matching~\cite{lipman2022flow}.

In particular, RF~\cite{liu2022rectified} connects the noise distribution $\pi_0$ and data distribution $\pi_1$ via a deterministic linear path. Given a data-noise pair $(\vx_0, \vx_1)$, the intermediate state at time $t \in [0,1]$ is defined as strict linear interpolation:
\begin{equation}
\label{eq:rf_interp}
    \vx_t = (1-t)\vx_0 + t\vx_1.
\end{equation}
RF learns a velocity field $p_{\vtheta}(\vx, t)$ to match the constant flow direction $\vx_1 - \vx_0$ by minimizing the following objective:
\begin{equation}
\label{eq:rf_objective}
    \mathcal{L}(\vtheta)
    = \mathbb{E}_{(\vx_0,\vx_1), t}
    \Big[ \big\| (\vx_1 - \vx_0) - p_{\vtheta}(\vx_t, t) \big\|_2^2 \Big].
\end{equation}
Sampling corresponds to solving the ODE $d\vz_t/dt = p_{\vtheta}(\vz_t, t)$ initialized from $\vz_0 \sim \pi_0$. Formally, RF discretizes the following integral of the learned velocity field over the time interval (via the Euler method, for instance):
\begin{equation}
\label{eq:rf_sampling}
    \vz_1 = \vz_0 + \int_0^1 p_{\vtheta}(\vz_t, t) \, dt.
\end{equation}

\subsection{Diffusion Language Models}

Discrete diffusion models~\cite{sohl2015deep,austin2021structured} are generative models tailored to discrete data, such as token sequences. Analogous to CDMs, they define a forward Markov process of discrete state that progressively corrupts the data toward a simple prior, typically a uniform distribution over tokens or an absorbing mask state. The model then learns the corresponding reverse dynamics to generate discrete data from this prior. A special case of discrete diffusion, masked diffusion models (MDMs) ~\cite{loudiscrete,shi2024simplified,sahoo2024simple,ou2024your}, have demonstrated strong potential and favorable scaling properties in language modeling.

\begin{figure*}[!t]
    \centering
\includegraphics[width=.95\linewidth]{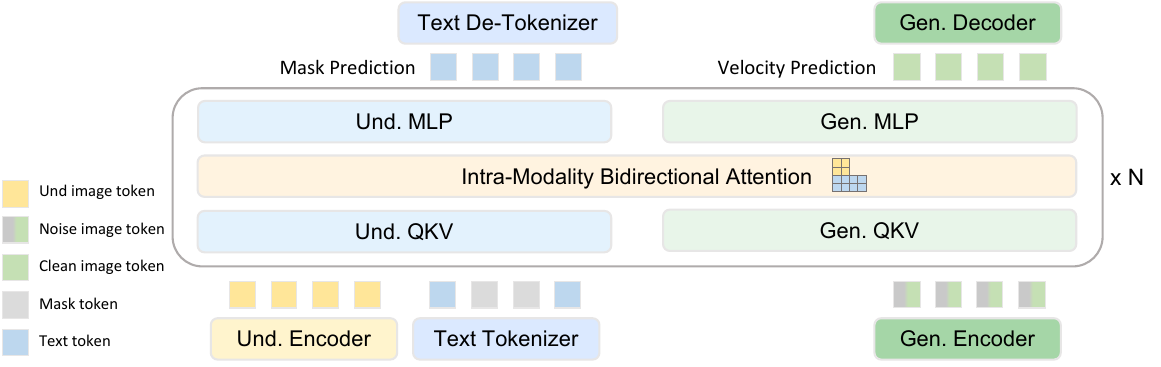}
    \caption{\textbf{Overview of LLaDA-o: the mixture of diffusion framework.}}
    \label{fig:architecture}
\end{figure*}

Formally, let $\vx_0 = [\vx^i]_{i=1}^N$ denote a sequence of $N$ tokens, and let $\text{[M]}$ represent a special mask token. In the forward process, tokens are independently corrupted based on a time step $t$ uniformly sampled from $[0,1]$. Let $\alpha_t$ be a continuous monotonically increasing function and $\alpha_0 = 0$ and $\alpha_1 =1$. Each token either remains unchanged with probability $\alpha_t$ or is replaced by $\text{[M]}$ with probability $1-\alpha_t$ as follows:
\begin{equation}
\label{eq:forward}
\begin{aligned}
    q_{t|0}(\vx_t|\vx_0) &= \prod_{i=0}^{N-1} q_{t|0}(\vx_t^i|\vx_0^i), \\
    q_{t|0}(\vx_t^i|\vx_0^i) &= \begin{cases}
                        \alpha_t, & \vx_t^i = \vx_0^i, \\
                        1-\alpha_t, & \vx_t^i = \text{[M]}.
                        \end{cases}
\end{aligned}
\end{equation}
The reverse process reconstructs the sequence iteratively. In particular, consider the transformation from time $t$ to $s$ (where $0 \le s < t \le 1$). Tokens that are already predicted remain fixed. Conversely, masked tokens either stay masked with probability $\frac{1-\alpha_s}{1-\alpha_t}$ or are decoded with probability $\frac{\alpha_s-\alpha_t}{1-\alpha_t}$ based on the model prediction $p_{\vtheta}(\vx_0^i|\vx_t)$:
\begin{equation}
\label{eq:reverse}
\begin{aligned}
    q_{s|t}(\vx_s|\vx_t) &= \prod_{i=0}^{N-1} q_{s|t}(\vx_s^i|\vx_t), \\
    q_{s|t}(\vx_s^i|\vx_t) &= 
        \begin{cases}
            1, & \vx_t^i \neq \text{[M]}, \vx_s^i = \vx_t^i,\\
            \frac{1-\alpha_s}{1-\alpha_t}, & \vx_t^i = \text{[M]}, \vx_s^i = \text{[M]}, \\
            \frac{\alpha_s-\alpha_t}{1-\alpha_t} p_{\vtheta}(\vx_0^i|\vx_t), & \vx_t^i =\text{[M]}, \vx_s^i \neq \text{[M]}, \\
            0, & \textrm{otherwise}.
        \end{cases}
\end{aligned}
\end{equation}
During inference, MDMs generate text by iteratively simulating this reverse transition, gradually converting a fully masked sequence into coherent text.

Intuitively, MDMs function as mask predictors, aiming to recover masked tokens from the observed context. Formally, the training objective of MDMs is defined as:
\begin{align}
\label{eq:training_objective}
\! \mathcal{L}(\vtheta) \!=\! \int_0^1 \! \frac{1}{t} \mathbb{E}_{\vx_0, \vx_t\sim q_{t|0}}\Big[\! \sum_{i : \vx_t^i = \text{[M]}} \! \! - \! \log p_{\vtheta}(\vx_0^i|\vx_t) \Big] d t.
\end{align}
Diffusion large language models (dLLMs)~\cite{nie2025large,ye2025dream} demonstrate the viability of masked diffusion at scale, achieving competitive performance while enabling parallel decoding and flexible generation control.

\section{Method}

Motivated by the discussion in Sec.~\ref{sec:intro}, \textbf{LLaDA-o} unifies multimodal understanding and generation within a single diffusion framework by separating discrete and continuous modalities, coupling them through a shared and efficient attention mechanism, and enabling adaptive-length training and inference, detailed as follows.

\subsection{The Mixture of Diffusion Framework}

At the core of LLaDA-o is a hybrid diffusion design that treats discrete tokens and continuous visual latents with their respective optimal parameterizations. 
However, naively co-training these modalities in a single dense model is often ineffective~\cite{li2025dual}: the two branches operate on heterogeneous state spaces and corruption processes, which can induce objective mismatch and gradient interference, leading to training conflicts and suboptimal performance.

To address these challenges, we propose \emph{Mixture of Diffusion} (MoD), a unified multi-modal diffusion framework illustrated in Fig.~\ref{fig:architecture}. Inspired by modality-factorized designs~\cite{liang2024mixture,deng2025emerging}, MoD employs two diffusion experts to decouple the processing of discrete and continuous modalities: an \emph{understanding expert} that handles text and visual encoder tokens via masked diffusion, and a \emph{generation expert} that handles visual latent tokens via continuous diffusion. While the experts allow for specialized processing, they share the same self-attention backbone (see Sec.~\ref{sec:imba}) to ensure effective cross-modality interaction.

Specifically, the understanding expert integrates a vision encoder~\cite{zhai2023sigmoid}, a lightweight two-layer MLP, and a diffusion language model~\cite{nie2025large} sequentially. 
The image is encoded into semantic visual tokens by the encoder, projected into the language token space by the MLP, and jointly processed with prompt tokens by the language model. Given a training sample $(\vv, \vp, \vr_0)$, where $\vv$ denotes the projected image tokens, $\vp$ denotes the prompt, and $\vr_0$ denotes the ground-truth response, the entire expert is trained to optimize the variant of Eq.~(\ref{eq:training_objective}) as follows: 
\begin{align}
\label{eq:visual_training_objective}
    \mathcal{L}_{\textrm{und}} = \int_0^1 \frac{1}{t} \mathbb{E}_{\substack{\vv,\vp \\ \vr_0,\vr_t}}\Bigg[ \sum_{\substack{i : \vr_t^i = \text{[M]}}} - \log p_{\vtheta}(\vr_0^i|\vv,\vp,\vr_t) \Bigg] d t.
\end{align}
In parallel, the generation expert comprises a variational autoencoder (VAE)~\cite{kingma2013auto} and a diffusion Transformer~\cite{peebles2023scalable}. 
The VAE maps between images and visual latent tokens and its parameters are kept frozen during training. 
Given the relevant variables $(\vp, \vv_0, \veps)$, where $\vp$ denotes the prompt, $\vv_0$ denotes the ground-truth image tokens from the VAE, and $\veps$ denotes a Gaussian noise, the training objective follows Eq.~(\ref{eq:rf_objective}):
\begin{equation}
\label{eq:gen_rf_objective}
    \mathcal{L}_{\textrm{gen}}
    = \mathbb{E}_{\vp, \veps,\vv_0, \vv_t,\,t}
    \Big[ \big\| (\vv_0 - \veps) - p_{\vtheta}(\vp,\vv_t, t) \big\|_2^2 \Big].
\end{equation}
It is worth noting that both the input image and text in the generation task are also processed by the understanding expert. As a result, the corresponding parameters are jointly trained. Detailed training protocols for multi-turn dialogue and interleaved data are provided in Appendix~\ref{app:multi_turn_dialog}.

\begin{figure*}[t!]
    \centering
    \begin{subfigure}[b]{0.22\textwidth}
        \centering
        \includegraphics[width=\textwidth]{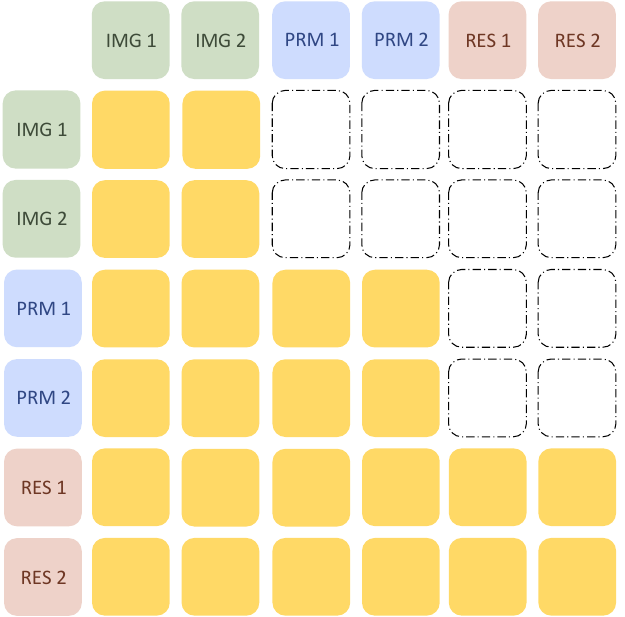}
        \caption{Multimodal Understanding}
        \label{fig:und_attn}
      \end{subfigure}
    \hfill
    \begin{subfigure}[b]{0.22\textwidth}
        \centering        \includegraphics[width=\textwidth]{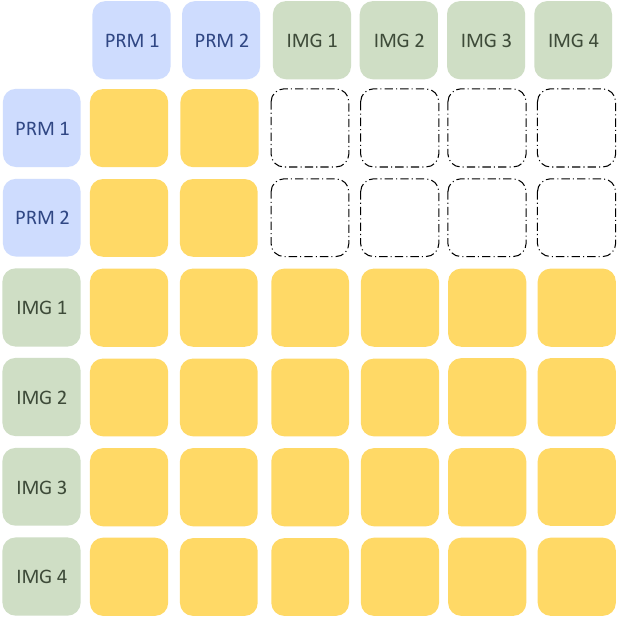}
        \caption{Multimodal Generation}
        \label{fig:t2i_attn}
      \end{subfigure}
    \hfill
    \begin{subfigure}[b]{0.22\textwidth}
        \centering
        \includegraphics[width=\textwidth]{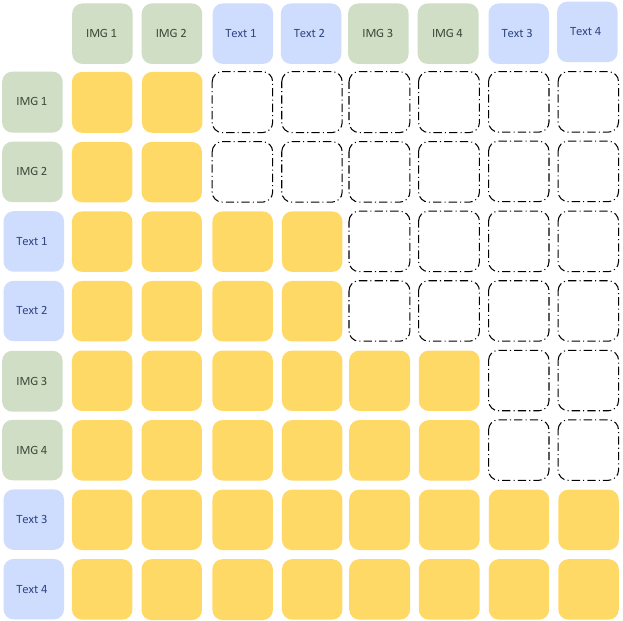}
        \caption{Interleaved Data}
        \label{fig:mix_mo_attn}
      \end{subfigure}
    \hfill
    \begin{subfigure}[b]{0.224\textwidth}
        \centering
        \includegraphics[width=\textwidth]{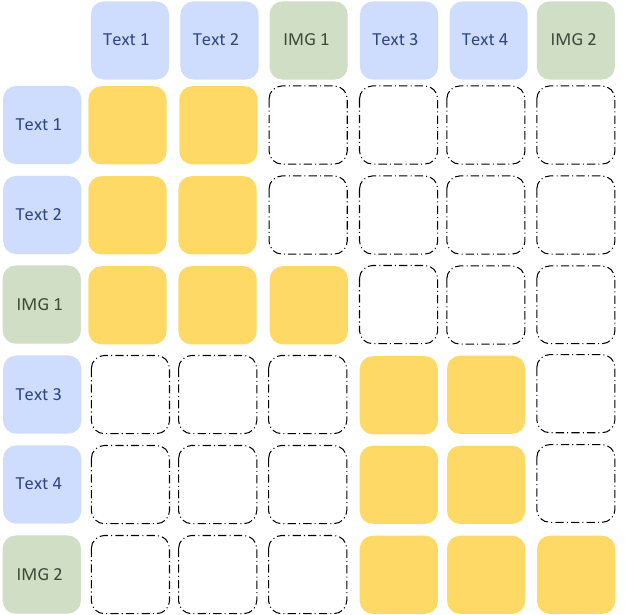}
        \caption{Cross-Sample Isolation}
        \label{fig:multi_sample_attn}
      \end{subfigure}
    \caption{\textbf{Implementation of intra-modality bidirectional attention.} Yellow blocks indicate \textit{unmasked} attention, while dashed white boxes denote \textit{masked} attention.  
Text sequences are explicitly partitioned into Prompts (\texttt{PRM}) and Responses (\texttt{RES}) in cases (a--b).}
\label{fig:attn_masks}
\end{figure*}

\subsection{Intra-Modality Bidirectional Attention}
\label{sec:imba}

Although MoD decouples modality-specific processing via different experts, they share a common attention backbone for cross-modality interaction. Using global attention in this setting is inefficient because it recomputes attention over the entire sequence at every denoising step, even when the condition (e.g., an input image or a text prompt) stays fixed. 

To this end, we propose \emph{intra-modality bidirectional attention}, a simple yet efficient attention scheme tailored for MoD. In particular, we partition an input sequence into modality blocks, apply full attention within each block, and enforce causal attention across blocks. This design preserves rich intra-modality context while enabling efficient inference: conditional blocks form a fixed prefix whose Key-Value (KV) cache can be reused across denoising steps, avoiding redundant computation on long sequences. 

The proposed attention scheme can be implemented by introducing appropriate attention masks between modality blocks. 
As illustrated in Fig.~\ref{fig:attn_masks}(a--c), this principle naturally accommodates different data types, including multimodal understanding, generation, and interleaved data. Besides, we use sample packing with strict isolation between samples (see Fig.~\ref{fig:multi_sample_attn}) to prevent cross-sample interference. 
 
Compared to a representative baseline~\cite{you2025llada} with globally bidirectional attention, our attention design achieves a 5.9 times speedup in practice (see Fig.~\ref{fig:mathvista_tradeoff}).

\subsection{Adaptive Length Augmentation}

The understanding expert adopts masked diffusion for text modeling, typically assuming a fixed target length at inference time. To enable flexible generation without introducing architectural changes or violating sample isolation, we design a data-centric strategy called \emph{adaptive length augmentation} that is fully compatible with our attention scheme.

As summarized in Alg.~\ref{alg:adalen}, during training, the target response in each individual sample is stochastically perturbed in two complementary ways. With probability $p_{\text{ext}}$, a bounded number of \texttt{[EOS]} tokens are appended to the original response, exposing the model to explicit termination at different positions. With probability $p_{\text{trunc}}$, the response is truncated to a random prefix, encouraging the model to learn proper continuation from partial targets. Importantly, both operations preserve strict sample isolation and do not require modifying the attention structure or sequence packing strategy.

At inference (see Alg.~\ref{alg:lladao_gen_text} in Appendix.~\ref{app:text_generation_process}), we perform block-wise generation~\cite{arriola2025block}  under intra-modality bidirectional attention: the fixed conditioning blocks (images and prompt) are cached once, and response tokens are generated iteratively by appending a length-$L$ masked block and denoising it. If \texttt{[EOS]} appears with high confidence, decoding terminates; otherwise, the completed block is cached and generation proceeds to the next block. This strategy enables efficient variable-length generation while fully reusing the KV cache of the fixed prefix.

Compared to prior approaches that rely on architectural modifications or multi-sample packing with globally bidirectional attention to handle variable-length outputs~\cite{Dreamon2025,kim2025any,yang2025diffusion}, our method remains lightweight, architecture-agnostic, and naturally compatible with sample isolation and efficient inference.

\begin{algorithm}
\caption{Training with Adaptive Length}
\label{alg:adalen}
\begin{algorithmic}[1]
    \STATE {\bfseries Input:} Model parameters $\boldsymbol{\theta}$; training data $(\vv,\vp,\vr_0)$; extension probability $p_{\text{ext}}$; truncation probability $p_{\text{trunc}}$.
    \REPEAT
        \STATE $\tilde{\vr}_0 \leftarrow \vr_0$, \ $u\sim\mathcal{U}(0,1)$
        \IF{$u < p_{\text{ext}}$}
            \STATE Sample an integer $k$ uniformly between $(1, |\vr_0|)$ and update $\tilde{\vr}_0$ by appending $k$ EOS tokens.
        \ELSIF{$u < p_{\text{ext}}+p_{\text{trunc}} \ \AND\ |\vr_0|>16$}
            \STATE Sample an integer $\ell$ uniformly between $(1, |\vr_0|-1)$ and update $\tilde{\vr}_0$ by truncating it to the first $\ell$ tokens.
        \ENDIF
        \STATE Train on $(\vv,\vp,\tilde{\vr}_0)$ with loss only on response tokens $\tilde{\vr}_0$; update $\boldsymbol{\theta}$
    \UNTIL{convergence}
\end{algorithmic}
\end{algorithm}

\begin{table*}[t!]
    \centering    
    \caption{\textbf{Evaluation on multimodal understanding benchmarks.} The symbol $^\dagger$ denotes results from LaViDa-O~\cite{li2025lavidao}, while $^\star$ indicates results we evaluated using official checkpoints and inference scripts. ``-'' represents missing data, and ``Diff.'' refers to diffusion language models. Notably, LLaDA-o achieves state-of-the-art performance among diffusion-based unified multimodal models.}
    \label{tab:multimodal_understanding_results}
    \begin{adjustbox}{max width=\textwidth}
    \setlength{\tabcolsep}{4.5pt} 
    \begin{tabular}{l|ccccccccccc}
      \toprule
      Model & MMMU & \multicolumn{2}{c}{MME} & SeedB & MMB & MathVerse & MathVista & AI2D & ChartQA & DocVQA & InfoVQA \\
      & val & cog. & perp. & image & en-dev & mini-vis. & testmini & - & - & val & val \\
      \midrule
      \rowcolor{gray!10} \multicolumn{12}{c}{AR Based} \\
      \midrule
      Emu3~\cite{wang2024emu3} & 31.6 & - & - & 68.2 & 58.5 & - & - & 70.0 & 68.6 & 76.3 & 43.8  \\
      Janus-Pro~\cite{chen2025janus} & 41.0 & - & 1567 & 72.1 & 79.2 & - & - & - & - & - & - \\
      MetaMorph~\cite{tong2025metamorph} & 41.8 & - & - & 71.8 & 75.2 & - & - & - & 37.1 & - & - \\
      Show-o~\cite{xie2024show} & 27.4 & - & 1232 & - & - & - & - & - & - & - & - \\
      Show-o2~\cite{xie2025show} & 48.9 & - & 1620 & - & 79.3 & - & - & 78.6 & - & - & - \\
      BAGEL~\cite{deng2025emerging} & \textbf{55.3} & - & \textbf{1687} & - & \textbf{85.0} & - & \textbf{73.1} & - & - & - & - \\
      \midrule
      \rowcolor{gray!10} \multicolumn{12}{c}{Diff. Based} \\
      \midrule
      LaViDa-L~\cite{li2025lavida} & 43.3 & 341 & 1365 & - & 70.5 & 27.2 & 44.8 & 70.0 & 64.6 & 59.0 & 34.2 \\
      Dimple~\cite{yu2025dimple} & 45.2 & 432 & 1514 & - & 74.6 & - & 42.3 & 74.4 & 63.4 & - & - \\
      LLaDA-V~\cite{you2025llada} & 48.6 & 491 & 1507 & 74.8 & 82.9 & 28.5 & 59.7 & 77.8 & 78.3 & 83.9 & \textbf{66.3} \\
      MMaDA~\cite{yang2025mmada} & 30.2 & 242$^\dagger$ & 1410 & 64.2 & 68.5 & 13.5$^\dagger$ & 33.7$^\dagger$ & 66.6$^\dagger$ & 9.8$^\dagger$ & 10.9$^\dagger$ & 14.9$^\dagger$ \\
      Lumina-DiMOO~\cite{xin2025lumina} & \textbf{58.6} & - & \textbf{1534} & \textbf{83.1} & \textbf{84.5} & 10.3$^\star$ & 30.3$^\star$ & 43.2$^\star$ & 8.3$^\star$ & 7.2$^\star$ & 6.2$^\star$ \\
      LaViDa-O~\cite{li2025lavidao} & 45.1 & 488 & 1431 & - & 76.4 & 36.9 & 56.9 & 76.7 & 80.0 & 73.7 & 44.6 \\
      \rowcolor{cyan!10} LLaDA-o & 44.9 & \textbf{549} & 1412 & 75.3 & 71.1 & \textbf{37.1} & \textbf{66.1} & \textbf{79.3} & \textbf{87.9} & \textbf{91.5} & 54.7 \\
      \bottomrule
    \end{tabular}
    \end{adjustbox}
\end{table*}

\section{Experiments}

We present experimental settings, results and analyses.

\subsection{Experimental Settings}
\label{sec:experimental_settings}
\textbf{Model.} For the understanding expert, we use the representative LLaDA-8B-Instruct~\cite{nie2025large} to initialize the language model. As the vision encoder, we use SigLIP~\cite{zhai2023sigmoid}, which has shown strong performance in many MLLMs, and adopt a randomly initialized two-layer MLP as the projector. For the generation expert, we use the same diffusion transformer architecture as the masked predictor architecture in LLaDA~\cite{nie2025large} and is initialized from it, while additional conditional parameters for time embeddings are randomly initialized. We use VAE of FLUX~\cite{flux2024} as the vision encoder for generation due to its strong reconstruction quality. 

\textbf{Training strategy.}
We train LLaDA-o in three stages to progressively scale both data difficulty and generation fidelity. In Stage~1, we use large-scale image understanding data together with image generation data, where generation is restricted to resolutions up to $512$ to stabilize training. In Stage~2, we incorporate multimodal reasoning data and reuse a high-quality subset of the Stage~1 image generation data, while increasing the generation resolution to $1024$ to improve high-resolution synthesis. Notably, we do not apply \emph{adaptive length augmentation} for multimodal understanding in the first two stages. In Stage~3, we jointly apply \emph{adaptive length augmentation} to activate variable-length generation for the understanding expert and add more high-quality image generation data, aligning the model with flexible text decoding and stronger visual generation. For more details on the training strategy and data, please refer to Appendix~\ref{app:training_strategy_data_settings}.

\textbf{Evaluation.} We evaluate LLaDA-o on a broad set of benchmarks to reflect the main requirements of unified multimodal models: general knowledge understanding, reasoning, and fine-grained perception, as well as image generation. For multimodal understanding, we cover multidisciplinary knowledge (MMMU~\cite{yue2024mmmu}, MME~\cite{fu2023mme}, SEED-Bench~\cite{li2023seed}, and MMBench~\cite{liu2024mmbench}), mathematical reasoning (MathVerse~\cite{zhang2024mathverse} and MathVista~\cite{lu2023mathvista}), and chart/document understanding (AI2D~\cite{kembhavi2016diagram}, ChartQA~\cite{masry2022chartqa}, DocVQA~\cite{mathew2021docvqa}, and InfoVQA~\cite{mathew2022infographicvqa}). For text-to-image generation, we use two widely used benchmarks that test complementary aspects of generation: GenEval~\cite{ghosh2023geneval}, which verifies fine-grained compositional attributes via an object-centric detection pipeline (e.g., object count, spatial relations, and color binding), and DPG-Bench~\cite{hu2024ella}, which evaluates faithful rendering of long, information-dense prompts with complex entity relationships and rich descriptions.

\subsection{Benchmark Results} 

\textbf{Multimodal understanding.} We compare LLaDA-o with unified multimodal models and multimodal large language models in Tab.~\ref{tab:multimodal_understanding_results}. Notably, LLaDA-o achieves \emph{state-of-the-art} performance among omni-diffusion models (e.g., LaViDa-O~\cite{li2025lavidao}), demonstrating the effectiveness of the Mixture of Diffusion. This advantage is particularly evident  on mathematical reasoning (e.g., MathVista~\cite{lu2023mathvista}) and chart/document understanding (e.g., ChartQA~\cite{masry2022chartqa}). These improvements support the effectiveness of our MoD framework. Compared with state-of-the-art autoregressive model BAGEL~\cite{deng2025emerging}, LLaDA-o is generally weaker, which is expected given BAGEL's stronger language backbone. BAGEL uses Qwen2.5-7B-Instruct~\cite{qwen2.5} trained on 18T tokens, while our LLaDA-8B-Instruct~\cite{nie2025large} is trained on 2.3T tokens, and this gap is reflected in language capability (e.g., 84.8 vs.\ 49.4 on HumanEval). Despite this disadvantage, LLaDA-o narrows the gap; for example, on MathVista it achieves 66.1, improving over LLaDA-V~\cite{you2025llada} (59.7) and approaching BAGEL (73.1). We believe MoD will further improve as masked diffusion backbones continue to improve.


\begin{table*}[t!]
  \centering
  \caption{\textbf{Evaluation of text-to-image generation on the GenEval benchmark}~\cite{ghosh2023geneval}. ``Gen.,'' ``Obj.,'' and ``Attr.'' denote generation, object, and attribute, respectively, while ``-'' indicates missing data. Compared to state-of-the-art unified multimodal models, LLaDA-o demonstrates superior performance, particularly in two-object and color-specific generation tasks. Following the protocols of BAGEL~\cite{deng2025emerging} and Show-o2~\cite{xie2025show}, we evaluate the results using rewritten prompts.}
  \label{tab:gen_eval_results}
  \begin{adjustbox}{max width=\textwidth}
  \begin{tabular}{cc|cccccc|c}
    \toprule
    Model & \# Params & Single Obj. & Two Obj. & Counting & Colors & Position & Color Attri. & Overall$\uparrow$ \\
    \midrule
    \rowcolor{gray!10} \multicolumn{9}{c}{Gen. Only} \\
    \midrule
    PixArt-${\alpha}$~\cite{chen2023pixart} & 0.6B & 0.98 & 0.50 & 0.44 & 0.80 & 0.08 & 0.07 & 0.48 \\
    DALL-E 3~\cite{betker2023improving} & - & 0.96 & 0.87 & 0.47 & 0.83 & \textbf{0.43} & 0.45 & 0.67 \\
    SD3-Medium~\cite{esser2024scaling} & 2B & \textbf{0.99} & \textbf{0.94} & 0.72 & \textbf{0.89} & 0.33 & \textbf{0.60} & \textbf{0.74}  \\
    FLUX.1-dev~\cite{flux2024} & 12B & 0.98 & 0.81 & \textbf{0.74} & 0.79 & 0.22 & 0.45 & 0.66  \\
    \midrule
    \rowcolor{gray!10} \multicolumn{9}{c}{Unified} \\
    \midrule
    Emu3~\cite{wang2024emu3} & 8B & - & - & - & - & - & - & 0.66  \\
    Janus~\cite{wu2025janus} & 1.3B & 0.97 & 0.68 & 0.30 & 0.84 & 0.46 & 0.42 & 0.61  \\
    Janus-Pro~\cite{chen2025janus} & 7B & 0.99 & 0.89 & 0.59 & 0.90 & 0.79 & 0.66 & 0.80   \\
    Mogao~\cite{liao2025mogao} & 7B & \textbf{1.00} & 0.97 & 0.83 & 0.93 & 0.84 & 0.80 & 0.89  \\
    MMaDA~\cite{yang2025mmada} & 8B & 0.99 & 0.76 & 0.61 & 0.84 & 0.20 & 0.37 & 0.63  \\
    Show-o~\cite{xie2024show} & 1.3B & 0.98 & 0.80 & 0.66 & 0.84 & 0.31 & 0.50 & 0.68 \\
    BAGEL~\cite{deng2025emerging} & 7B MoT & 0.98 & 0.95 & 0.84 & 0.95 & 0.78 & 0.77 & 0.88 \\
    LaViDa-O~\cite{li2025lavidao} & - & - & - & - & - & - & - & 0.89  \\
    Lumina-DiMOO~\cite{xin2025lumina} & 8B & \textbf{1.00} & 0.96 & \textbf{0.87} & 0.95 & \textbf{0.85} & 0.82 & \textbf{0.91}  \\
    Show-o2~\cite{xie2025show} & 7B & \textbf{1.00} & 0.87 & 0.58 & 0.92 & 0.52 & 0.62 & 0.76  \\
    \rowcolor{cyan!10}
    LLaDA-o (ours) & 8B MoT & 0.99 & \textbf{0.98} & 0.73 & \textbf{0.96} & 0.69 & \textbf{0.83} & 0.86  \\
    \bottomrule
  \end{tabular}
  \end{adjustbox}
\end{table*}

\begin{table*}[t!]
  \centering
  \caption{\textbf{Evaluation of text-to-image generation on DPG-Bench}~\cite{hu2024ella}. The symbol $^\dagger$ denotes results from Lumina-DiMOO~\cite{xin2025lumina}, while ``Gen.'' stands for generation and ``-'' indicates missing data. Notably, LLaDA-o achieves state-of-the-art performance compared to previous generation-only and unified models.}
  \label{tab:dpg_bench_results}
  \begin{adjustbox}{max width=\textwidth}
  \begin{tabular}{cc|ccccc|c}
    \toprule
    Model & \# Params & Global & Entity & Attribute & Relation & Other & Overall$\uparrow$ \\
    \midrule
    \rowcolor{gray!10} \multicolumn{8}{c}{Gen. Only} \\
    \midrule
    PixArt-${\alpha}$~\cite{chen2023pixart} & 0.6B & 74.97 & 79.32 & 78.60 & 82.57 & 76.96 & 71.11 \\
    PixArt-${\Sigma}$~\cite{chen2024pixart} & 0.6B & 86.89 & 82.89 & 88.94 & 86.59 & 87.68 & 80.54 \\
    DALL-E 3~\cite{betker2023improving} & - & \textbf{90.97} & 89.61 & 88.39 & 90.58 & \textbf{89.83} & 83.50 \\
    SD3-Medium~\cite{esser2024scaling} & 2B & 87.90 & \textbf{91.01} & 88.83 & 80.70 & 88.68 & \textbf{84.08}  \\
    FLUX.1-dev~\cite{flux2024} & 12B & 74.35 & 90.00 & \textbf{88.96} & \textbf{90.87} & 88.33 & 83.84  \\
    \midrule
    \rowcolor{gray!10} \multicolumn{8}{c}{Unified} \\
    \midrule
    Emu3-DPO~\cite{wang2024emu3} & 8B & - & - & - & - & - & 81.60  \\
    Janus~\cite{wu2025janus} & 1.3B & 82.33 & 87.38 & 87.70 & 85.46 & 86.41 & 79.68  \\
    Janus-Pro~\cite{chen2025janus} & 7B & 86.90 & 88.90 & 89.40 & 89.32 & 89.48 & 84.19   \\
    Mogao~\cite{liao2025mogao} & 7B & 82.37 & 90.03 & 88.26 & 93.18 & 85.40 & 84.33  \\
    MMaDA$^\dagger$~\cite{yang2025mmada} & 8B & 77.81 & 78.48 & 81.74 & 84.79 & 63.20 & 69.97  \\
    Show-o~\cite{xie2024show} & 1.3B & - & - & - & - & - & 67.48 \\
    BAGEL~\cite{deng2025emerging} & 7B MoT & 88.94 & 90.37 & \textbf{91.29} & 90.82 & 88.67 & 85.07 \\
    LaViDa-O~\cite{li2025lavidao} & - & - & - & - & - & - & 83.20  \\
    Lumina-DiMOO~\cite{xin2025lumina} & 8B & 81.46 & 92.08 & 88.98 & \textbf{94.31} & 82.00 & 86.04  \\
    Show-o2~\cite{xie2025show} & 7B & 89.00 & 91.78 & 89.96 & 91.81 & 91.64 & 86.14  \\
    \rowcolor{cyan!10}
    LLaDA-o (ours) & 8B MoT & \textbf{92.91} & \textbf{93.30} & 90.40 & 91.75 & \textbf{92.79} & \textbf{87.04}  \\
    \bottomrule
  \end{tabular}
  \end{adjustbox}
\end{table*}

\textbf{Text-to-image generation.} We evaluate LLaDA-o against state-of-the-art generation-only models and unified multimodal models on GenEval~\cite{ghosh2023geneval} and DPG-Bench~\cite{hu2024ella} (Tabs.~\ref{tab:gen_eval_results} and~\ref{tab:dpg_bench_results}). On GenEval, LLaDA-o outperforms most strong models such as Janus-Pro~\cite{chen2025janus} and SD3-Medium~\cite{esser2024scaling}. Although it is slightly behind Lumina-DiMOO~\cite{xin2025lumina} and Mogao~\cite{liao2025mogao} overall, it performs better on two-object generation and color binding. Notably, on DPG-Bench, LLaDA-o achieves state-of-the-art performance (87.04), surpassing Show-o2~\cite{xie2025show} and Lumina-DiMOO, indicating strong generation quality for long, information-dense prompts. These results support the effectiveness of our MoD framework in combining continuous diffusion model with a dLLM-based backbone for unified multimodal generation. Qualitatively, Fig.~\ref{fig:generated_image_demo} shows that LLaDA-o produces more visually appealing images with richer fine-grained details than Lumina-DiMOO and LaViDa-O~\cite{li2025lavidao} while following the instructions.

We also provide additional qualitative text-to-image samples in Appendix.~\ref{app:addtional_generated_image}.

\begin{table}[t!]
\centering
\caption{\textbf{Effect of confidence threshold on MathVista.} We report the accuracy (\%) and throughput (tokens/s) across varying confidence thresholds, with the block length fixed at 64.}
\label{tab:confidence_abla}
\begin{adjustbox}{max width=\linewidth}
\begin{tabular}{lcccccc}
\toprule
\textbf{Confidence Threshold ($r$)} & \textbf{0.2} & \textbf{0.4} & \textbf{0.6} & \textbf{0.8} & \textbf{0.9} & \textbf{1.0} \\
\midrule
\textbf{Accuracy (\%)} & 57.8 & 62.4 & 64.2 &  65.0 & 65.9 & 65.8 \\
\textbf{Throughput (tokens/s)} & 203.9 & 134.5 & 97.3 & 64.1 & 52.2 & 24.3 \\
\bottomrule
\end{tabular}
\end{adjustbox}
\end{table}

\begin{figure}[!t]
    \centering
    \includegraphics[width=\linewidth]{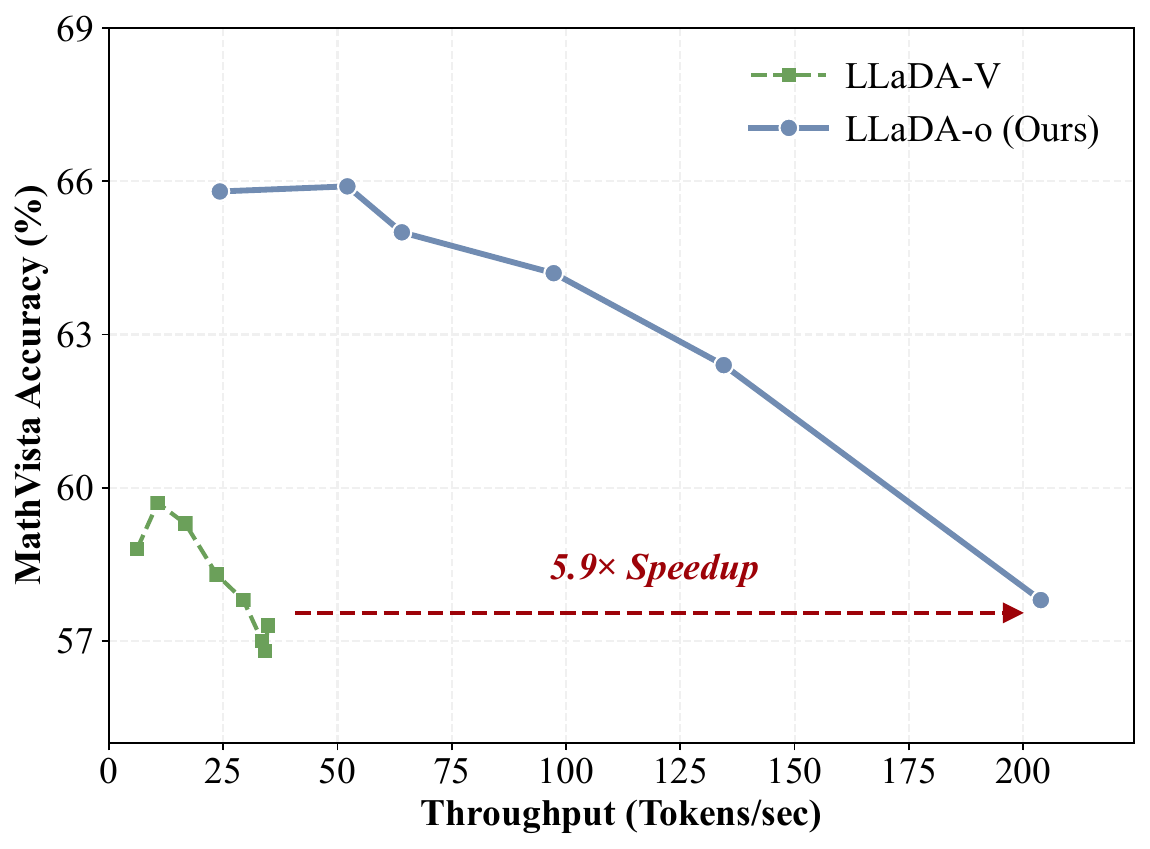}    
    \caption{\textbf{Comparison of inference efficiency on MathVista.} We visualize the throughput-accuracy trade-off by varying the confidence threshold for LLaDA-o and the refresh interval ($n$) of Fast-dLLM applied to LLaDA-V. Our approach outperforms LLaDA-V, achieving a $5.9\times$ speedup with comparable performance.}
    \label{fig:mathvista_tradeoff}
\end{figure}

\subsection{Further Analysis}
Tab.~\ref{tab:confidence_abla} and Fig.~\ref{fig:mathvista_tradeoff} analyze the inference efficiency and generation quality. Unlike autoregressive models, LLaDA-o offers the flexibility to regulate this trade-off via the confidence threshold: raising the threshold prioritizes accuracy by selecting only high-confidence tokens, while lowering it accelerates generation. Empirically, a threshold of 0.9 achieves the optimal balance. When comparing with state-of-the-art baselines on MathVista (Fig.~\ref{fig:mathvista_tradeoff}), LLaDA-o demonstrates significant efficiency gains. Most notably, LLaDA-o delivers a $5.9\times$ speedup compared to LLaDA-V. This substantial improvement validates the effectiveness of our \emph{intra-modality bidirectional attention}, which reduces computational redundancy and enables efficient inference.


We provide comprehensive analysis of the variable-length behavior in Appendix~\ref{app:comparison_lladav_mismatch_block},\ref{app:block_to_variable_effect} and Tab.~\ref{tab:block_length_abla}. Qualitatively, unlike LLaDA-V, which produces redundant or incomplete text depending on mismatched block settings, LLaDA-o generates content of appropriate length consistent across varying block sizes ($L \in \{16, \dots, 128\}$). Quantitatively, Tab.~\ref{tab:block_length_abla} reveals that the generated length remains relatively stable: increasing the block length from 32 to 96 results in only a minor decrease in average token count (165 to 145) while improving accuracy (63.6\% to 66.2\%). These results demonstrate that the output length is mainly driven by the input content rather than the preset block size, confirming the effectiveness of our \emph{adaptive length augmentation}.

\begin{table}[t!]
\centering
\caption{\textbf{Effect of block length on MathVista.} We report the average number of generated tokens under different block lengths, with the confidence threshold fixed at 0.95.}
\label{tab:block_length_abla}
\begin{adjustbox}{max width=\linewidth}
\begin{tabular}{lcccc}
\toprule
\textbf{Block length} & \textbf{32} & \textbf{64} & \textbf{96} & \textbf{128} \\
\midrule
\textbf{Average tokens} & 165 & 148 & 145 &  154  \\
\textbf{Accuracy (\%)} & 63.6 & 66.1 & 66.2 & 65.3  \\
\bottomrule
\end{tabular}
\end{adjustbox}
\end{table}

Finally, Tab.~\ref{tab:stage_ablation} studies the effect of each training stage on text-to-image generation. Results on GenEval and DPG-Bench improve from Stage~1 to Stage~3, with Stage~3 performing best (0.82 and 87.0), supporting the effectiveness of our multi-stage training pipeline. For completeness, we report the computational cost of our three-stage training pipeline in Appendix~\ref{app:computation_cost}.

\begin{table}[t!]
\centering
\caption{\textbf{Text-to-image generation performance across training stages.} We report the evaluation results on GenEval and DPG-Bench to demonstrate the performance progression. In this table, we use the original prompts of GenEval.}
\label{tab:stage_ablation}
\begin{adjustbox}{max width=0.75\linewidth}
\begin{tabular}{c|cc}
\toprule
\textbf{Training Stage} & \textbf{GenEval} & \textbf{DPG-Bench} \\
\midrule
Stage 1 & 0.74 & 86.1 \\
Stage 2 & 0.78 & 86.2 \\
Stage 3 & \textbf{0.82} & \textbf{87.0} \\
\bottomrule
\end{tabular}
\end{adjustbox}
\end{table}

\section{Conclusion}

We presented LLaDA-o, a length-adaptive omni diffusion model for multimodal understanding and generation. Built on a Mixture of Diffusion framework with a shared efficient attention backbone and a data-centric adaptive length training strategy, LLaDA-o enables stable multimodal training and flexible-length generation. Experimental results demonstrate strong performance on multimodal understanding and text-to-image generation tasks. We believe that as masked diffusion models continue to advance in language modeling, LLaDA-o potentially provides a promising foundation for future omni diffusion approaches.


\bibliography{main}
\bibliographystyle{icml2026}

\newpage
\appendix
\onecolumn

\section{Related Work}

\textbf{Diffusion large language models.} 
Recently, diffusion large language models (dLLMs)~\cite{nie2025large,ye2025dream,zhu2025lladamoe} have emerged based on masked diffusion models (MDMs)~\cite{ou2024your,loudiscrete,shi2024simplified,sahoo2024simple,you2025effective}, which are a special case of discrete diffusion models~\cite{sohl2015deep,hoogeboom2021argmax}. Through large-scale pretraining and supervised fine-tuning, these models have achieved performance comparable to strong autoregressive models such as LLaMA3. This demonstrates the practical applicability and strong potential of dLLMs as alternatives to ARMs. Beyond text generation, dLLMs have also made remarkable progress in various domains, including multimodal understanding~\cite{you2025llada,yu2025dimple,li2025lavida}, audio understanding~\cite{zhou2025diffa}, reinforcement Learning~\cite{zhu2025llada,zhao2025d1,ou2025principled} and vision-language-action tasks~\cite{wen2025llada,wen2025dvla,liang2025discrete}. Most relevant to our work are unified multimodal models~\cite{li2025lavidao,xin2025lumina,yang2025mmada}. However, unlike these approaches that rely on masked diffusion models, we employ continuous diffusion models for image generation, thereby avoiding the information loss caused by image discretization.

\textbf{Unified multimodal models} are initially dominated by autoregressive architectures~\cite{team2024chameleon,wang2024emu3}, which leverage structural consistency with LLMs via unified next-token prediction. Subsequently, hybrid approaches combine autoregressive text processing with diffusion-based image generation~\cite{zhou2024transfusion,xie2024show,tong2025metamorph,liao2025mogao,deng2025emerging}. Concurrently, Diffusion Large Language Models (dLLMs)~\cite{nie2025large,ye2025dream,zhu2025lladamoe} have achieved performance comparable to strong autoregressive models via large-scale pretraining and supervised fine-tuning. Building on this success, recent dLLM-based unified multimodal models~\cite{li2025lavidao,xin2025lumina,yang2025mmada,swerdlow2025unified} employ masked diffusion mechanisms for both multimodal understanding and generation.

\section{Experimental Details}

\subsection{Training Stages and Data Setup}
\label{app:training_strategy_data_settings}
The training of LLaDA-o proceeds in a preliminary alignment phase followed by three progressive stages.

\textbf{Projector Alignment.}
Prior to the main training, we focus on aligning visual representations with the understanding expert. We train the MLP projector using the Stage 1 data from Infinity-MM~\cite{gu2024infinity}, while keeping all other components frozen.

\textbf{Stage 1: Foundation Setup.}
In this stage, we establish the baseline capabilities using large-scale image understanding and generation data. We restrict image generation to a resolution of $512$ and do not apply \emph{adaptive length augmentation} for multimodal understanding.
For \textit{text}, we use an in-house 10M SFT dataset.
For \textit{multimodal understanding}, we utilize Stage 2-4 data from Infinity-MM~\cite{gu2024infinity}, MAmmoTH-VL-Instruct-12M~\cite{guo2025mammoth}, LLaVA-OneVision-1.5~\cite{an2025llava}, and FineVision~\cite{an2025llava}.
For \textit{text-to-image generation}, we combine image captioning data (from Infinity-MM~\cite{gu2024infinity} and LLaVA-OneVision-1.5~\cite{an2025llava}) with generation datasets including PD12M~\cite{meyer2024public}, Text-to-Image-2M~\cite{t2i2m}, BLIP3o-Pretrain~\cite{chen2025blip3}, Nexus-Gen~\cite{zhang2025nexus}, FLUX-Reason-6M~\cite{fang2025flux}, and synthetic data from Qwen-Image~\cite{wu2025qwen}.

\textbf{Stage 2: High-Resolution and Reasoning.}
We further incorporate multimodal reasoning data and increase the generation resolution to $1024$. \emph{Adaptive length augmentation} remains disabled for understanding tasks.
In terms of data, we switch the \textit{understanding} source to Honey-Data-15M~\cite{zhang2025bee}.
For \textit{generation}, we refine the dataset by removing image captions and PD12M~\cite{meyer2024public}, while increasing the sampling ratio of the remaining high-quality data.
Additionally, we introduce \textit{interleaved multimodal data} from X2Edit~\cite{ma2025x2edit} and OmniGen2~\cite{wu2025omnigen2}, along with editing data from ShareGPT-4o-Image~\cite{chen2025sharegpt} and OpenGPT-4o-Image~\cite{chen2025opengpt}. The text data remains unchanged.

\textbf{Stage 3: Variable-Length Refinement.}
In the final stage, we jointly apply \emph{adaptive length augmentation} to activate variable-length generation for the understanding expert and fine-tune the model with high-quality data.
For \textit{understanding}, we incorporate Honey-Data-1M and retain Honey-Data-15M~\cite{zhang2025bee} with a reduced ratio.
For \textit{generation}, we add premium datasets including ShareGPT-4o-Image~\cite{chen2025sharegpt}, OpenGPT-4o-Image~\cite{chen2025opengpt}, BLIP3o-60k~\cite{chen2025blip3}, and GenRef~\cite{zhuo2025reflection}, while reducing the ratio of the Stage 2 generation data.
For \textit{interleaved data}, we remove X2Edit~\cite{ma2025x2edit} and keep the rest consistent with Stage 2.

\subsection{Multi-turn Dialogue Data and Interleaved Multimodal Data Setup}
\label{app:multi_turn_dialog}
For multi-turn dialogues, LLaDA~\cite{nie2025large} randomly selects one turn for training. It concatenates the preceding dialogue history, including both prompts and responses, as the input context for that turn, and computes the loss only on the selected turn's response. This setup is also used in LaViDa~\cite{li2025lavida} and Dimple~\cite{yu2025dimple}. In contrast, LLaDA-V~\cite{you2025llada} computes the loss on the response of every turn in the dialogue. In LLaDA-o, we follow the LLaDA-V strategy, since it achieves strong performance and shows data scalability in LLaDA-V.

For interleaved multimodal data, we treat the text only as a condition (i.e., prompt) for image generation. Thus, we compute the continuous diffusion loss only on each turn of image tokens and mask out the loss on text tokens, preventing the image objective from interfering with text representations.

\subsection{Prompts of Selected Generated Images}
\label{app:prompt_for_generated}
Tab.~\ref{tab:image_prompts} lists the text prompts used to generate the samples shown in Fig.~\ref{fig:generated_image_demo}. The Image IDs in the table correspond to the spatial arrangement of the images: IDs 1--5 represent the top row (from left to right), and IDs 6--10 represent the bottom row.

\begin{table}[h!]
\centering
\caption{\textbf{Prompts for the generated samples.} The Image IDs correspond to the order of images in Fig.~\ref{fig:generated_image_demo}, arranged from left to right and top to bottom.}
\label{tab:image_prompts}
\begin{adjustbox}{max width=\linewidth}
\setlength{\tabcolsep}{8pt}       
\renewcommand{\arraystretch}{1.2} 
\begin{tabular}{c p{0.85\linewidth}}
\toprule
\textbf{Image ID} & \textbf{Prompt} \\
\midrule
1 & Bright scene, aerial view, ancient city, fantasy, gorgeous light, mirror reflection, high detail, wide angle lens. \\
2 & A photorealistic corgi sitting calmly in a traditional Chinese tea house, steam from the teacup forming a tiny dragon, morning light, detailed fur. \\
3 & A detective cat in a trench coat standing in a rainy alleyway, cyberpunk neon signs reflecting in puddles, film noir style, volumetric fog, dramatic shadows, cinematic shot. \\
4 & An astronaut riding a horse on the moon, oil painting by Van Gogh. \\
5 & A miniature library inside a vintage lightbulb, warm cozy light. \\
6 & A raccoon wearing a tiny backpack and using a map to navigate a misty mountain trail, storybook illustration, soft colors. \\
7 & A complex landscape of a futuristic city made entirely of folded origami paper, soft warm paper texture, subsurface scattering, dramatic lighting from inside the paper buildings, tilt-shift photography. \\
8 & A transparent glass piano in a dark forest, glowing bioluminescent mushrooms growing inside the piano, magical dust floating in the air, fantasy illustration, intricate details. \\
9 & A red sports car made entirely of knitted wool, soft studio lighting. \\
10 & Portrait of a futuristic robot, pencil sketch by Leonardo da Vinci. \\
\bottomrule
\end{tabular}
\end{adjustbox}
\end{table}

\subsection{Computational Cost}
\label{app:computation_cost}
Tab.~\ref{tab:gpu_resources} details the computational resources and time required for each training stage of LLaDA-o. The primary training phases (Stage 1 and Stage 2) are conducted on 256 NVIDIA H800 GPUs, accounting for the majority of the computational cost. The final refinement stage (Stage 3) is performed using 64 NVIDIA A100 GPUs.

\begin{table}[t!]
\centering
\caption{\textbf{Computational resources and training cost.} The table details the GPU hardware and total GPU hours consumed in each training stage.}
\label{tab:gpu_resources}
\begin{adjustbox}{max width=0.75\linewidth}
\begin{tabular}{c|cc}
\toprule
\textbf{Training Stage} & \textbf{GPUs} & \textbf{GPU Hours} \\
\midrule
Stage 1 & 256 H800 & 55,296 \\
Stage 2 & 256 H800 & 30,720 \\
Stage 3 & ~64 A100 & ~1,536 \\
\bottomrule
\end{tabular}
\end{adjustbox}
\end{table}

\subsection{Text Generation Process of LLaDA-o}
\label{app:text_generation_process}
During inference, we adopt a blockwise sampling procedure~\cite{arriola2025block}. As outlined in Algorithm~\ref{alg:lladao_gen_text}, we first cache the fixed prefix (images and prompt). Subsequently, we extend the sequence by appending a block of length $L$, initialized with mask tokens. Within each block, we perform iterative denoising: tokens with prediction confidence exceeding a threshold $\tau$ are accepted, while others remain masked for the next step. If an \texttt{[EOS]} token is detected, we truncate the sequence and terminate; otherwise, we update the cache with the completed block and proceed to the next masked block.

\begin{algorithm}[h!]
\caption{Text Generation Process of LLaDA-o.}
\label{alg:lladao_gen_text}
\begin{algorithmic}[1]
    \STATE {\bfseries Input:} Parameters $\boldsymbol{\theta}$; Images $\mathcal{V}$; Prompt $\mathcal{P}$; Block size $L$; Threshold $\tau$.
    \STATE {\bfseries Output:} Generated sequence $\mathbf{y}$.
    
    \STATE $\mathcal{C} \leftarrow \textsc{Encode}(\mathcal{V}, \mathcal{P}; \boldsymbol{\theta}); \ \mathbf{y} \leftarrow \emptyset$ \hfill \COMMENT{Initial Context Encoding}
    
    \LOOP
        \STATE $\mathbf{b} \leftarrow [\texttt{MASK}]^L$; \ $\mathcal{M} \leftarrow \{1, \dots, L\}$ \hfill \COMMENT{Init Block \& Mask indices}
        
        \WHILE{$\mathcal{M} \neq \emptyset$}
            \STATE $\mathbf{p} \leftarrow f_{\boldsymbol{\theta}}(\mathbf{b}, \mathcal{C})$ \hfill \COMMENT{Forward pass using parameters $\boldsymbol{\theta}$}
            \FOR{$i \in \mathcal{M}$}
                \IF{$\max(\mathbf{p}_i) > \tau$}
                    \STATE $\mathbf{b}_i \leftarrow \arg\max(\mathbf{p}_i)$ \hfill \COMMENT{Keep high-confidence tokens}
                    \STATE $\mathcal{M} \leftarrow \mathcal{M} \setminus \{i\}$
                \ENDIF
            \ENDFOR
        \ENDWHILE
        
        \IF{$\texttt{EOS} \in \mathbf{b}$}
            \STATE \textbf{return} $\mathbf{y} \parallel \textsc{Trunc}(\mathbf{b}, \texttt{EOS})$
        \ENDIF
        
        \STATE $\mathbf{y} \leftarrow \mathbf{y} \parallel \mathbf{b}; \ \mathcal{C} \leftarrow \textsc{UpdateCache}(\mathcal{C}, \mathbf{b})$
    \ENDLOOP
\end{algorithmic}
\end{algorithm}

\section{Additional Results}
\label{app:additional_results}
\subsection{Additional Generated Images}
\label{app:addtional_generated_image}
\begin{figure}[t!]
    \centering
    \includegraphics[width=\linewidth]{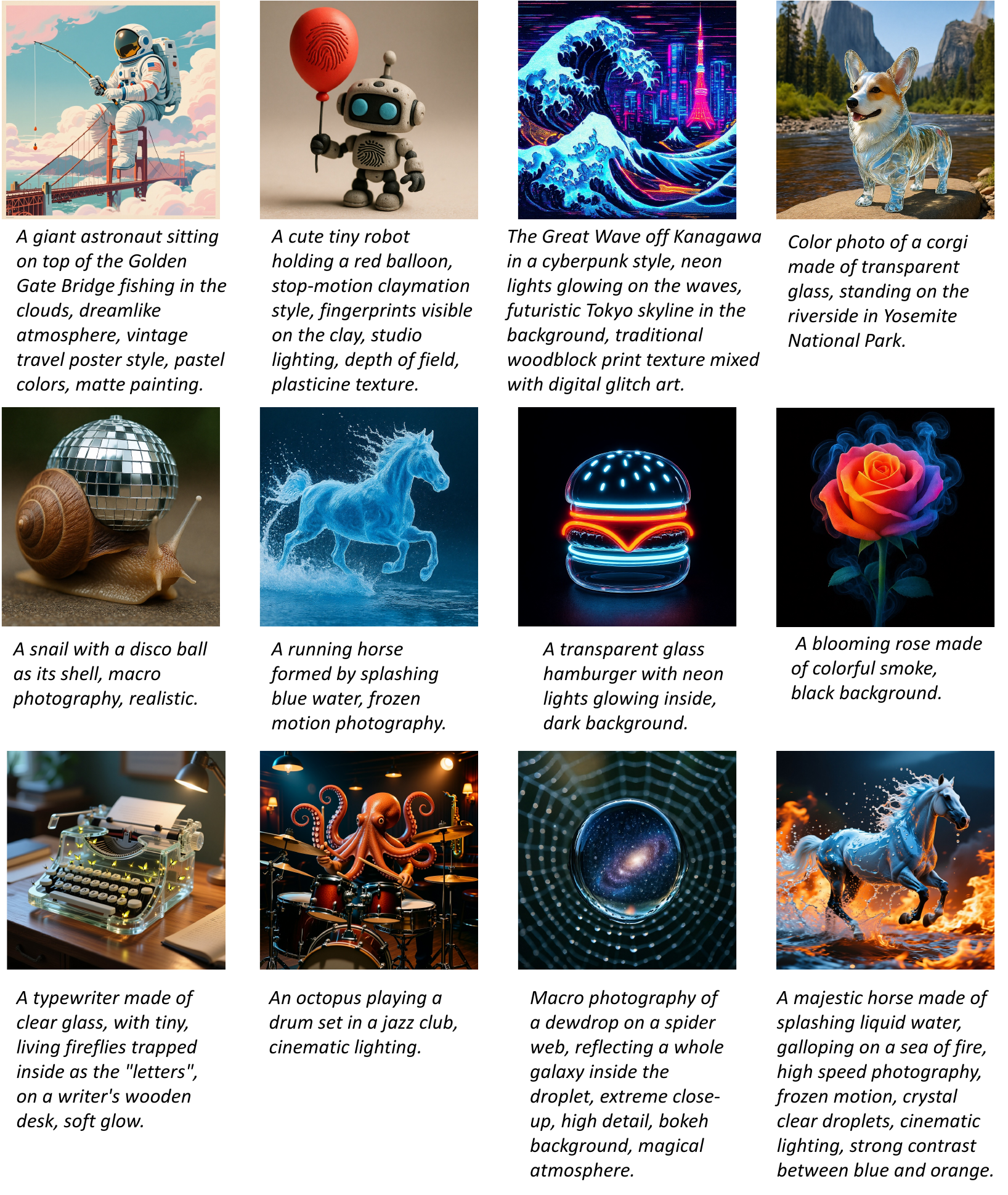}    
    \caption{\textbf{Additional generated samples.} We present 12 randomly selected images generated by LLaDA-o. For each sample, the prompt used for generation is shown below the corresponding image. All results are produced under the same setting as in the main paper.}
    \label{fig:app_generated_image}
\end{figure}

We provide additional qualitative examples generated by LLaDA-o in Fig.~\ref{fig:app_generated_image}. These samples further demonstrate the model's capability to produce high-quality images that are semantically aligned with user prompts, exhibiting both high fidelity and diversity.

\subsection{Qualitative Comparison with LLaDA-V under Mismatched Block Lengths}
\label{app:comparison_lladav_mismatch_block}

In this section, we present qualitative samples demonstrating the variable-length text generation capability of LLaDA-o. As shown in Tab.~\ref{tab:var_length_demo}, we compare LLaDA-o with the state-of-the-art LLaDA-V in scenarios where the semantic requirement of the prompt conflicts with the pre-defined generation block length.

First, as shown in the top row of Tab.~\ref{tab:var_length_demo}, when the user requests simple text extraction but provides a long block length (i.e., $L=64$), LLaDA-V tends to fill the entire window with redundant content. In contrast, LLaDA-o accurately extracts the text and correctly terminates, adhering to the user's intent. Second, as shown in the bottom row, when the user requests a detailed image description but assigns a short block length (i.e., $L=16$), LLaDA-V is constrained by the fixed window, resulting in an overly brief response. Conversely, LLaDA-o automatically extends the generation by appending additional blocks if no End-of-Sequence (EOS) token is detected within the current block, continuing until the generation is complete.

\begin{table*}[t!]
    \centering
    \caption{\textbf{Qualitative samples demonstrating variable-length generation.} \textbf{Top:} Under a long block length setting ($L=64$), LLaDA-V generates redundant content for a simple prompt, while LLaDA-o correctly terminates. \textbf{Bottom:} Under a short block length setting ($L=16$), LLaDA-V produces an overly brief response for a detailed prompt, whereas LLaDA-o dynamically adapts the output length to the user prompt.}
    \label{tab:var_length_demo}
    \scalebox{1}{
    \begin{adjustbox}{max width=\textwidth}
    \renewcommand{\arraystretch}{1.2} 
    \begin{tabular}{l p{15cm} }
    \toprule
    ~~~~User & What's written on this image, just give me the answer. \\ 
     & \includegraphics[width=.62\textwidth,valign=t]{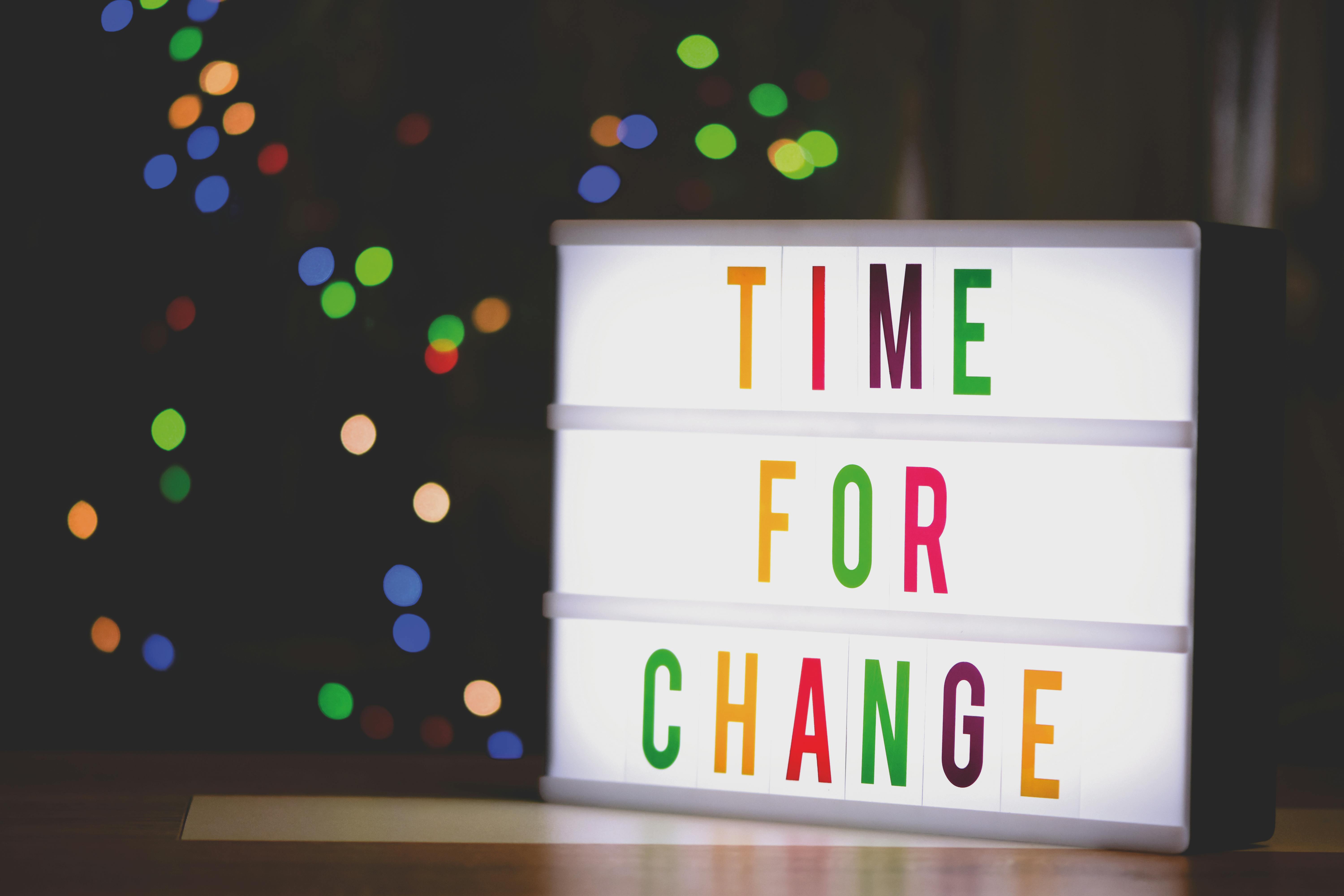}\\ 
    ~~~~LLaDA-V & The image displays a lightboard sign with the phrase  ``TIME FOR CHANGE'' written in bold, colorful letters. The sign is set against a backdrop of blurred lights, suggesting a festive or celebratory atmosphere. The text on the sign reads ``TIME FOR CHANGE,'' indicating a call to action or a message for transformation. \\
    ~~~~LLaDA-o & TIME FOR CHANGE \\
    \midrule
    ~~~~User & Please describe this image in detail. \\ 
     & \includegraphics[width=.62\textwidth,valign=t]{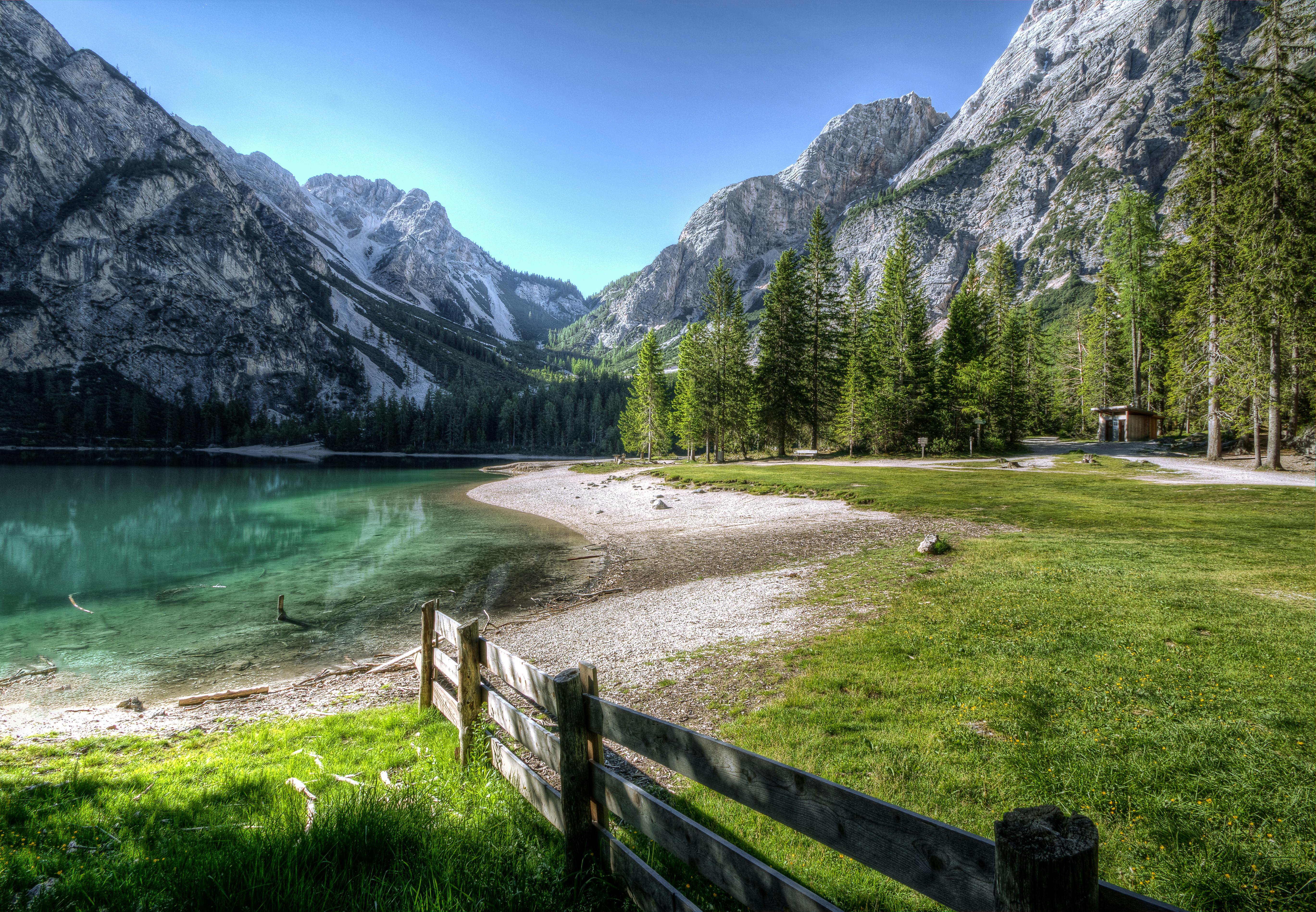}\\ 
    ~~~~LLaDA-V & The image captures a serene landscape featuring a tranquil lake surrounded by towering mountains. \\
    ~~~~LLaDA-o & The image captures a serene mountain landscape, likely a popular tourist destination. Dominating the scene is a large lake, its surface a vibrant shade of green, reflecting the surrounding mountains. The lake's edge is lined with a wooden fence, adding a rustic charm to the scene. Beyond the fence, a grassy area stretches out, dotted with trees and a small building, possibly a shelter or a visitor center. The building is nestled among the trees, blending seamlessly with the natural surroundings. The mountains rise majestically in the background, their peaks dusted with snow, suggesting a high-altitude location. The sky above is a clear blue, adding to the overall tranquility of the scene. The image does not contain any visible text. \\
    \bottomrule
    \end{tabular}
    \end{adjustbox}
    }
\end{table*}

\subsection{Effect of Block Length on Variable-Length Generation}
\label{app:block_to_variable_effect}

In this section, we present qualitative samples to illustrate the variable-length generation behavior of LLaDA-o. As shown in Tab.~\ref{tab:change_block_demo}, we maintain the same image and prompt while varying the pre-defined generation block length $L \in \{16, 32, 64, 128\}$. Across these different settings, the generated outputs remain largely consistent, with only minor fluctuations in response length. This suggests that LLaDA-o is not rigidly constrained by the specific block length and can dynamically adapt the effective output length based on the user prompt and image content.

\begin{table*}[t!]
    \centering
    \caption{\textbf{Qualitative samples demonstrating variable-length generation.} We show generations from LLaDA-o for the same image and the same prompt under different block length settings ($L\in\{16,32,64,128\}$). Despite changing $L$, the output length remains largely stable, indicating that LLaDA-o can dynamically adapt the output length to the user prompt and image.}
    \label{tab:change_block_demo}
    \scalebox{1}{
    \begin{adjustbox}{max width=\textwidth}
    \renewcommand{\arraystretch}{1.2} 
    \begin{tabular}{p{2.4cm} p{15cm} }
    \toprule
    ~~User & Please describe this image in detail. \\ 
     & \includegraphics[width=.62\textwidth,valign=t]{figs/und_4.jpg}\\ 
    LLaDA-o (block:16) & The image captures a serene mountain landscape, likely a popular tourist destination. Dominating the scene is a large lake, its surface a vibrant shade of green, reflecting the surrounding mountains. The lake's edge is lined with a wooden fence, adding a rustic charm to the scene. Beyond the fence, a grassy area stretches out, dotted with trees and a small building, possibly a shelter or a visitor center. The building is nestled among the trees, blending seamlessly with the natural surroundings. The mountains rise majestically in the background, their peaks dusted with snow, suggesting a high-altitude location. The sky above is a clear blue, adding to the overall tranquility of the scene. The image does not contain any visible text. \\
    \midrule
    LLaDA-o (block:32) & The image depicts a serene, picturesque natural landscape featuring a tranquil lake surrounded by rugged mountains and lush greenery. The lake is a vibrant shade of turquoise, reflecting the clear blue sky above. A wooden fence runs along the edge of the lake, separating the grassy area from the water. The fence is simple and rustic, with vertical wooden posts and horizontal rails. In the foreground, there is a well-maintained grassy area with a wooden fence. The grass is lush and green, indicating a healthy, natural environment. A few yellow flowers are scattered throughout the grass, adding a touch of color to the scene. The overall atmosphere of the image is peaceful and inviting, suggesting a perfect spot for relaxation and enjoyment of the beauty of nature. \\
    \midrule
    LLaDA-o (block:64) & The image depicts a serene natural landscape featuring a small, clear lake surrounded by lush green grass and tall evergreen trees. The lake is nestled in a valley, with rugged, rocky mountains rising in the background. The sky is clear and blue, indicating a sunny day. In the foreground, there is a wooden fence that runs along the edge of the grassy area, separating it from the lake. A small wooden structure, possibly a cabin or a small house, is visible on the right side of the image, partially obscured by the tall trees. The overall scene is peaceful and idyllic.\\
    \midrule
    LLaDA-o (block:128) & The image captures a serene mountain landscape, featuring a small, clear lake nestled in a valley. The lake is surrounded by lush green grass and trees, creating a stark contrast with the rocky mountains in the background. The mountains, with their rugged peaks, rise majestically against the clear blue sky. The perspective of the image is from a low angle, looking up at the mountains, which adds a sense of grandeur and majesty to the scene. The overall mood of the image is peaceful and tranquil, inviting the viewer to immerse themselves in the natural beauty of this mountainous paradise. \\
    \bottomrule
    \end{tabular}
    \end{adjustbox}
    }
\end{table*}


\end{document}